\def\year{2020}\relax
\documentclass[letterpaper]{article} 
\usepackage{aaai20}  
\usepackage{times}  
\usepackage{helvet} 
\usepackage{courier}  
\usepackage[hyphens]{url}  
\usepackage{graphicx} 
\usepackage{graphicx}  
\usepackage{url}
\usepackage{multirow}
\usepackage{graphicx}
\usepackage{amsmath}
\usepackage{amssymb}
\usepackage{subfig}
\title{Effective Modeling of Encoder-Decoder Architecture \\ for Joint Entity and Relation Extraction}

\author{Tapas Nayak \and Hwee Tou Ng \\
 Department of Computer Science \\
 National University of Singapore \\
 nayakt@u.nus.edu, nght@comp.nus.edu.sg
}
 
\begin{document}

\maketitle

\begin{abstract}
A relation tuple consists of two entities and the relation between them, and often such tuples are found in unstructured text. There may be multiple relation tuples present in a text and they may share one or both entities among them. Extracting such relation tuples from a sentence is a difficult task and sharing of entities or overlapping entities among the tuples makes it more challenging. Most prior work adopted a pipeline approach where entities were identified first followed by finding the relations among them, thus missing the interaction among the relation tuples in a sentence. In this paper, we propose two approaches to use encoder-decoder architecture for jointly extracting entities and relations. In the first approach, we propose a representation scheme for relation tuples which enables the decoder to generate one word at a time like machine translation models and still finds all the tuples present in a sentence with full entity names of different length and with overlapping entities. Next, we propose a pointer network-based decoding approach where an entire tuple is generated at every time step. Experiments on the publicly available New York Times corpus show that our proposed approaches outperform previous work and achieve significantly higher F1 scores.
\end{abstract}

\section{Introduction}

\begin{table*}[ht]
\centering
\resizebox{1.8\columnwidth}{!}{
\begin{tabular}{l|l|l}
\hline
Source sentence & \multicolumn{2}{l}{\begin{tabular}[c]{@{}l@{}}Anti-Ethiopia riots erupted in Mogadishu , the capital of Somalia , on Friday , while masked gunmen emerged \\for the first time on the streets , a day after Ethiopian-backed troops captured the city from Islamist forces .\end{tabular}} \\ \hline
Target: word-based decoding  & \multicolumn{2}{l}{\begin{tabular}[c]{@{}l@{}}Somalia ; Mogadishu ; /location/country/capital $\vert$ Somalia ; Mogadishu ; /location/location/contains \end{tabular}}                                                        \\ \hline
Target: pointer network-based decoding & \multicolumn{2}{l}{\begin{tabular}[c]{@{}l@{}}$<$9 9 4 4 /location/country/capital$>$ $<$9 9 4 4 /location/location/contains$>$ \end{tabular}}                                                        \\ \hline
\end{tabular}
}
\caption{Relation tuple representation for encoder-decoder models.}
\label{tab:mt_schemes}
\end{table*}

\noindent Distantly-supervised information extraction systems extract relation tuples with a set of pre-defined relations from text. Traditionally, researchers \cite{mintz2009distant,riedel2010modeling,hoffmann2011knowledge} use pipeline approaches where a named entity recognition (NER) system is used to identify the entities in a sentence and then a classifier is used to find the relation (or no relation) between them. However, due to the complete separation of entity detection and relation classification, these models miss the interaction between multiple relation tuples present in a sentence. 

Recently, several neural network-based models \cite{katiyar2016investigating,miwa2016end} were proposed to jointly extract entities and relations from a sentence. These models used a parameter-sharing mechanism to extract the entities and relations in the same network. But they still find the relations after identifying all the entities and do not fully capture the interaction among multiple tuples. \citeauthor{zheng2017joint} (\citeyear{zheng2017joint}) proposed a joint extraction model based on neural sequence tagging scheme. But their model could not extract tuples with overlapping entities in a sentence as it could not assign more than one tag to a word. \citeauthor{zeng2018copyre} (\citeyear{zeng2018copyre}) proposed a neural encoder-decoder model for extracting relation tuples with overlapping entities. However, they used a copy mechanism to copy only the last token of the entities, thus this model could not extract the full entity names. Also, their best performing model used a separate decoder to extract each tuple which limited the power of their model. This model was trained with a fixed number of decoders and could not extract tuples beyond that number during inference. Encoder-decoder models are powerful models and they are successful in many NLP tasks such as machine translation, sentence generation from structured data, and open information extraction.

In this paper, we explore how encoder-decoder models can be used effectively for extracting relation tuples from sentences. There are three major challenges in this task: (i) The model should be able to extract entities and relations together. (ii) It should be able to extract multiple tuples with overlapping entities. (iii) It should be able to extract exactly two entities of a tuple with their full names. To address these challenges, we propose two {\it novel} approaches using encoder-decoder architecture\footnote{The code and data of this paper can be found at https://github.com/nusnlp/PtrNetDecoding4JERE}. We first propose a new representation scheme for relation tuples (Table \ref{tab:mt_schemes}) such that it can represent multiple tuples with overlapping entities and different lengths of entities in a simple way. We employ an encoder-decoder model where the decoder extracts one word at a time like machine translation models. At the end of sequence generation, due to the unique representation of the tuples, we can extract the tuples from the sequence of words. Although this model performs quite well, generating one word at a time is somewhat unnatural for this task. Each tuple has exactly two entities and one relation, and each entity appears as a continuous text span in a sentence. The most effective way to identify them is to find their start and end location in the sentence. Each relation tuple can then be represented using five items: start and end location of the two entities and the relation between them (see Table \ref{tab:mt_schemes}). Keeping this in mind, we propose a pointer network-based decoding framework. This decoder consists of two pointer networks which find the start and end location of the two entities in a sentence, and a classification network which identifies the relation between them. At every time step of the decoding, this decoder extracts an entire relation tuple, not just a word. Experiments on the New York Times (NYT) datasets show that our approaches work effectively for this task and achieve state-of-the-art performance. To summarize, the contributions of this paper are as follows:

\noindent(1) We propose a new representation scheme for relation tuples such that an encoder-decoder model, which extracts one word at each time step, can still find multiple tuples with overlapping entities and tuples with multi-token entities from sentences. We also propose a masking-based copy mechanism to extract the entities from the source sentence only.\\
\noindent(2) We propose a modification in the decoding framework with pointer networks to make the encoder-decoder model more suitable for this task. At every time step, this decoder extracts an entire relation tuple, not just a word. This new decoding framework helps in speeding up the training process and uses less resources (GPU memory). This will be an important factor when we move from sentence-level tuple extraction to document-level extraction. \\
\noindent(3) Experiments on the NYT datasets show that our approaches outperform all the previous state-of-the-art models significantly and set a new benchmark on these datasets.

\section{Task Description}

A relation tuple consists of two entities and a relation. Such tuples can be found in sentences where an entity is a text span in a sentence and a relation comes from a pre-defined set $R$. These tuples may share one or both entities among them. Based on this, we divide the sentences into three classes: (i) {\em No Entity Overlap (NEO)}: A sentence in this class has one or more tuples, but they do not share any entities. (ii) {\em Entity Pair Overlap (EPO)}: A sentence in this class has more than one tuple, and at least two tuples share both the entities in the same or reverse order. (iii) {\em Single Entity Overlap (SEO)}: A sentence in this class has more than one tuple and at least two tuples share exactly one entity. It should be noted that a sentence can belong to both EPO and SEO classes. Our task is to extract all relation tuples present in a sentence.

\section{Encoder-Decoder Architecture}

In this task, input to the system is a sequence of words, and output is a set of relation tuples. In our first approach, we represent each tuple as {\em entity1 ; entity2 ; relation}. We use `;' as a separator token to separate the tuple components. Multiple tuples are separated using the `$\vert$' token. We have included one example of such representation in Table \ref{tab:mt_schemes}. Multiple relation tuples with overlapping entities and different lengths of entities can be represented in a simple way using these special tokens (; and $\vert$). During inference, after the end of sequence generation, relation tuples can be extracted easily using these special tokens. Due to this uniform representation scheme, where entity tokens, relation tokens, and special tokens are treated similarly, we use a shared vocabulary between the encoder and decoder which includes all of these tokens. The input sentence contains clue words for every relation which can help generate the relation tokens. We use two special tokens so that the model can distinguish between the beginning of a relation tuple and the beginning of a tuple component. To extract the relation tuples from a sentence using the encoder-decoder model, the model has to generate the entity tokens, find relation clue words and map them to the relation tokens, and generate the special tokens at appropriate time. Our experiments show that the encoder-decoder models can achieve this quite effectively.

\subsection{Embedding Layer \& Encoder}

We create a single vocabulary $V$ consisting of the source sentence tokens, relation names from relation set $R$, special separator tokens (`;', `$\vert$'), start-of-target-sequence token ({\em SOS}), end-of-target-sequence token ({\em EOS}), and unknown word token ({\em UNK}). Word-level embeddings are formed by two components: (1) pre-trained word vectors (2) character embedding-based feature vectors. We use a word embedding layer $\mathbf{E}_w \in \mathbb{R}^{\vert V \vert \times d_w}$ and a character embedding layer $\mathbf{E}_c \in \mathbb{R}^{\vert A \vert \times d_c}$, where $d_w$ is the dimension of word vectors, $A$ is the character alphabet of input sentence tokens, and $d_c$ is the dimension of character embedding vectors. Following \citeauthor{chiu2016named} (\citeyear{chiu2016named}), we use a convolutional neural network with max-pooling to extract a feature vector of size $d_f$ for every word. Word embeddings and character embedding-based feature vectors are concatenated ($\Vert$) to obtain the representation of the input tokens.

 A source sentence $\mathbf{S}$ is represented by vectors of its tokens $\mathbf{x}_1, \mathbf{x}_2,....,\mathbf{x}_n$, where $\mathbf{x}_i \in \mathbb{R}^{(d_w+d_f)}$ is the vector representation of the $i$th word and $n$ is the length of $\mathbf{S}$. These vectors $\mathbf{x}_i$ are passed to a bi-directional LSTM \cite{hochreiter1997long} (Bi-LSTM) to obtain the hidden representation $\mathbf{h}_i^E$. We set the hidden dimension of the forward and backward LSTM of the Bi-LSTM to be $d_h/2$ to obtain $\mathbf{h}_i^E \in \mathbb{R}^{d_h}$, where $d_h$ is the hidden dimension of the sequence generator LSTM of the decoder described below.

\subsection{Word-level Decoder \& Copy Mechanism}

A target sequence $\mathbf{T}$ is represented by only word embedding vectors of its tokens $\mathbf{y}_0, \mathbf{y}_1,....,\mathbf{y}_m$ where $\mathbf{y}_i \in \mathbb{R}^{d_w}$ is the embedding vector of the $i$th token and $m$ is the length of the target sequence. $\mathbf{y}_0$ and $\mathbf{y}_m$ represent the embedding vector of the {\em SOS} and {\em EOS} token respectively. The decoder generates one token at a time and stops when {\em EOS} is generated. We use an LSTM as the decoder and at time step $t$, the decoder takes the source sentence encoding ($\mathbf{e}_t \in \mathbb{R}^{d_h}$) and the previous target word embedding ($\mathbf{y}_{t-1}$) as the input and generates the hidden representation of the current token ($\mathbf{h}_t^D \in \mathbb{R}^{d_h}$). The sentence encoding vector $\mathbf{e}_t$ can be obtained using attention mechanism. $\mathbf{h}_t^D$ is projected to the vocabulary $V$ using a linear layer with weight matrix $\mathbf{W}_v \in \mathbb{R}^{\vert V \vert \times d_h}$ and bias vector $\mathbf{b}_v \in \mathbb{R}^{\vert V \vert}$ (projection layer). 
\begin{align*}
&\mathbf{h}_t^D = \mathrm{LSTM}(\mathbf{e}_t \Vert \mathbf{y}_{t-1}, \mathbf{h}_{t-1}^D)\\
&\hat{\mathbf{o}}_t = \mathbf{W}_v \mathbf{h}_t^D + \mathbf{b}_v, \quad
\mathbf{o}_t = \mathrm{softmax}(\hat{\mathbf{o}}_t)
\end{align*}
$\mathbf{o}_t$ represents the normalized scores of all the words in the embedding vocabulary at time step $t$. $\mathbf{h}_{t-1}^D$ is the previous hidden state of the LSTM.

The projection layer of the decoder maps the decoder output to the entire vocabulary. During training, we use the gold label target tokens directly. However, during inference, the decoder may predict a token from the vocabulary which is not present in the current sentence or the set of relations or the special tokens. To prevent this, we use a masking technique while applying the softmax operation at the projection layer. We mask (exclude) all words of the vocabulary except the current source sentence tokens, relation tokens, separator tokens (`;', `$\vert$'), {\em UNK}, and {\em EOS} tokens in the softmax operation. To mask (exclude) some word from softmax, we set the corresponding value in $\hat{\mathbf{o}}_t$ at $-\infty$ and the corresponding softmax score will be zero. This ensures the copying of entities from the source sentence only. We include the {\em UNK} token in the softmax operation to make sure that the model generates new entities during inference. If the decoder predicts an {\em UNK} token, we replace it with the corresponding source word which has the highest attention score. During inference, after decoding is finished, we extract all tuples based on the special tokens, remove duplicate tuples and tuples in which both entities are the same or tuples where the relation token is not from the relation set. This model is referred to as {\em WordDecoding} (WDec) henceforth.

\subsection{Pointer Network-Based Decoder}

In the second approach, we identify the entities in the sentence using their start and end locations. We remove the special tokens and relation names from the word vocabulary and word embeddings are used only at the encoder side along with character embeddings. We use an additional relation embedding matrix $\mathbf{E}_r \in \mathbb{R}^{\vert R \vert \times d_r}$ at the decoder side of our model, where  $R$ is the set of relations and $d_r$ is the dimension of relation vectors. The relation set $R$ includes a special relation token {\em EOS} which indicates the end of the sequence. Relation tuples are represented as a sequence $T=y_0, y_1,....,y_m$, where $y_t$ is a tuple consisting of four indexes in the source sentence indicating the start and end location of the two entities and a relation between them (see Table \ref{tab:mt_schemes}). $y_0$ is a dummy tuple that represents the start tuple of the sequence and $y_m$ functions as the end tuple of the sequence which has {\em EOS} as the relation (entities are ignored for this tuple). The decoder consists of an LSTM with hidden dimension $d_h$ to generate the sequence of tuples, two pointer networks to find the two entities, and a classification network to find the relation of a tuple. At time step $t$, the decoder takes the source sentence encoding ($\mathbf{e}_t \in \mathbb{R}^{d_h}$) and the representation of all previously generated tuples ($\mathbf{y}_{prev}=\sum_{j=0}^{t-1}\mathbf{y}_{j}$) as the input and generates the hidden representation of the current tuple, $\mathbf{h}_t^D \in \mathbb{R}^{d_h}$. The sentence encoding vector $\mathbf{e}_t$ is obtained using an attention mechanism as explained later. Relation tuples are a set and to prevent the decoder from generating the same tuple again, we pass the information about all previously generated tuples at each time step of decoding. $\mathbf{y}_j$ is the vector representation of the tuple predicted at time step $j < t$ and we use the zero vector ($\mathbf{y}_0=\overrightarrow{0}$) to represent the dummy tuple $y_0$. $\mathbf{h}_{t-1}^D$ is the hidden state of the LSTM at time step $t-1$.
\begin{align*}
&\mathbf{y}_{prev} = \sum_{j=0}^{t-1} \mathbf{y}_j, \quad
\mathbf{h}_t^D = \mathrm{LSTM}(\mathbf{e}_t \Vert \mathbf{y}_{prev}, \mathbf{h}_{t-1}^D)
\end{align*}

\begin{figure}[t]
\centering
\includegraphics[scale=0.33]{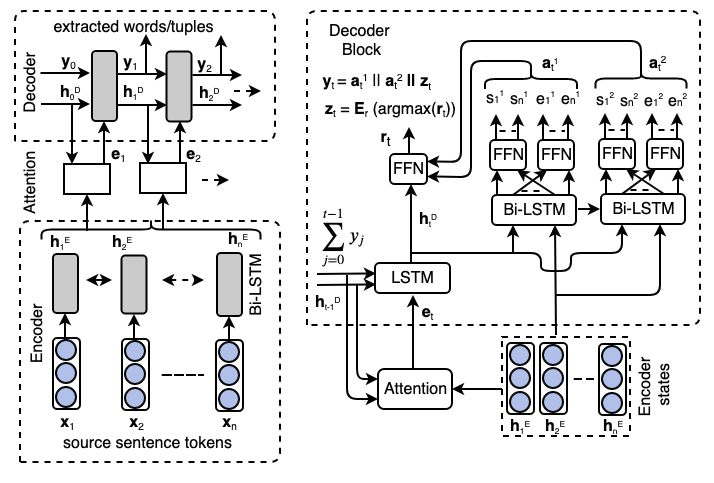}
\caption{The architecture of an encoder-decoder model (left) and a pointer network-based decoder block (right).}
\label{fig:model_diagram}
\end{figure}

\subsection{Relation Tuple Extraction}

After obtaining the hidden representation of the current tuple $\mathbf{h}_t^D$, we first find the start and end pointers of the two entities in the source sentence. We concatenate the vector $\mathbf{h}_t^D$ with the hidden vectors $\mathbf{h}_i^E$ of the encoder and pass them to a Bi-LSTM layer with hidden dimension $d_p$ for forward and backward LSTM. The hidden vectors of this Bi-LSTM layer $\mathbf{h}_i^k \in \mathbb{R}^{2d_p}$ are passed to two feed-forward networks (FFN) with softmax to convert each hidden vector into two scalar values between $0$ and $1$. Softmax operation is applied across all the words in the input sentence. These two scalar values represent the probability of the corresponding source sentence token to be the start and end location of the first entity. This Bi-LSTM layer with the two feed-forward layers is the first pointer network which identifies the first entity of the current relation tuple.
\begin{align*}
&\hat{s}_i^1 = \mathbf{W}_s^1 \mathbf{h}_i^k + {b}_s^1,
\quad \hat{e}_i^1 = \mathbf{W}_e^1 \mathbf{h}_i^k + {b}_e^1\\
&\mathbf{s}^1 = \mathrm{softmax}(\hat{\mathbf{s}}^1),
\quad \mathbf{e}^1 = \mathrm{softmax}(\hat{\mathbf{e}}^1)
\end{align*}

\noindent where $\mathbf{W}_s^1 \in \mathbb{R}^{1 \times 2d_p}$, $\mathbf{W}_e^1 \in \mathbb{R}^{1 \times 2d_p}$, ${b}_s^1$, and ${b}_e^1$ are the weights and bias parameters of the feed-forward layers. ${s}_i^1$, ${e}_i^1$ represent the normalized probabilities of the $i$th source word being the start and end token of the first entity of the predicted tuple. We use another pointer network to extract the second entity of the tuple. We concatenate the hidden vectors $\mathbf{h}_i^k$ with $\mathbf{h}_t^D$ and $\mathbf{h}_i^E$ and pass them to the second pointer network to obtain ${s}_i^2$ and ${e}_i^2$, which represent the normalized probabilities of the $i$th source word being the start and end of the second entity. These normalized probabilities are used to find the vector representation of the two entities, $\mathbf{a}_t^1$ and $\mathbf{a}_t^2$.
\begin{align*}
&\mathbf{a}_t^1 = \sum_{i=1}^n {s}_i^1 \mathbf{h}_i^k \Vert \sum_{i=1}^n {e}_i^1 \mathbf{h}_i^k, \quad 
\mathbf{a}_t^2 = \sum_{i=1}^n {s}_i^2 \mathbf{h}_i^l \Vert \sum_{i=1}^n {e}_i^2 \mathbf{h}_i^l
\end{align*}
We concatenate the entity vector representations $\mathbf{a}_t^1$ and $\mathbf{a}_t^2$ with $\mathbf{h}_t^D$ and pass it to a feed-forward network (FFN) with softmax to find the relation. This feed-forward layer has a weight matrix $\mathbf{W}_r \in \mathbb{R}^{\vert R \vert \times (8d_p + d_h)}$ and a bias vector $\mathbf{b}_r \in \mathbb{R}^{\vert R \vert}$.
\begin{align*}
&\mathbf{r}_t = \mathrm{softmax}(\mathbf{W}_r (\mathbf{a}_t^1 \Vert \mathbf{a}_t^2 \Vert  \mathbf{h}_t^D) + \mathbf{b}_r)\\
&\mathbf{z}_t=\mathbf{E}_r(\mathrm{argmax}(\mathbf{r}_t)), \quad
\mathbf{y}_t=\mathbf{a}_t^1 \Vert \mathbf{a}_t^2 \Vert \mathbf{z}_t
\end{align*}
$\mathbf{r}_t$ represents the normalized probabilities of the relation at time step $t$. The relation embedding vector $\mathbf{z}_t$ is obtained using $\mathrm{argmax}$ of $\mathbf{r}_t$ and $\mathbf{E}_r$. $\mathbf{y}_t \in \mathbb{R}^{(8d_p + d_r)}$ is the vector representation of the tuple predicted at time step $t$. During training, we pass the embedding vector of the gold label relation in place of the predicted relation. So the $\mathrm{argmax}$ function does not affect the back-propagation during training. The decoder stops the sequence generation process when the predicted relation is {\em EOS}. This is the classification network of the decoder.

During inference, we select the start and end location of the two entities such that the product of the four pointer probabilities is maximized keeping the constraints that the two entities do not overlap with each other and $1 \leq b \leq e \leq n$ where $b$ and $e$  are the start and end location of the corresponding entities. We first choose the start and end location of entity 1 based on the maximum product of the corresponding start and end pointer probabilities. Then we find entity 2 in a similar way excluding the span of entity 1 to avoid overlap. The same procedure is repeated but this time we first find entity 2 followed by entity 1. We choose that pair of entities which gives the higher product of four pointer probabilities between these two choices. This model is referred to as {\em PtrNetDecoding} (PNDec) henceforth.

\subsection{Attention Modeling}

We experimented with three different attention mechanisms for our word-level decoding model to obtain the source context vector $\mathbf{e}_t$:

\noindent(1) Avg.: The context vector is obtained by averaging the hidden vectors of the encoder:
       $\mathbf{e}_t=\frac{1}{n}\sum_{i=1}^n \mathbf{h}_i^E$ \\
\noindent(2)  N-gram: The context vector is obtained by the N-gram attention mechanism of \citeauthor{trisedya-etal-2019-neural} (\citeyear{trisedya-etal-2019-neural}) with N=3. \\ \\
        $\textnormal{a}_i^g=(\mathbf{h}_n^{E})^T \mathbf{V}^g \mathbf{w}_i^g$, \quad $\boldsymbol{\alpha}^g = \mathrm{softmax}(\mathbf{a}^g)$ \\ \\
        $\mathbf{e}_t=[\mathbf{h}_n^E \Vert \sum_{g=1}^N \mathbf{W}^g (\sum_{i=1}^{\vert G^g \vert} \alpha_i^g \mathbf{w}_i^g)$]\\ \\
        Here, $\mathbf{h}_n^E$ is the last hidden state of the encoder, $g \in \{1, 2, 3\}$ refers to the word gram combination, $G^g$ is the sequence of g-gram word representations for the input sentence, $\mathbf{w}_i^g$ is the $i$th g-gram vector (2-gram and 3-gram representations are obtained by average pooling), $\alpha_i^g$ is the normalized attention score for the $i$th g-gram vector, $\mathbf{W} \in \mathbb{R}^{d_h \times d_h}$ and $\mathbf{V} \in \mathbb{R}^{d_h \times d_h}$ are trainable parameters. \\
\noindent(3)  Single: The context vector is obtained by the attention mechanism proposed by \citeauthor{Bahdanau2014NeuralMT} (\citeyear{Bahdanau2014NeuralMT}). This attention mechanism gives the best performance with the word-level decoding model.  \\ \\
        $\mathbf{u}_t^i = \mathbf{W}_{u} \mathbf{h}_i^E, \quad \mathbf{q}_t^i = \mathbf{W}_{q} \mathbf{h}_{t-1}^D  + \mathbf{b}_{q}$,\\ \\
$\textnormal{a}_t^i = \mathbf{v}_a \tanh(\mathbf{q}_t^i + \mathbf{u}_t^i),
\quad \boldsymbol{\alpha}_t = \mathrm{softmax}(\mathbf{a}_t)$,\\ \\
$\mathbf{e}_t = \sum_{i=1}^n \alpha_t^i \mathbf{h}_i^E$ \\ \\
where $\mathbf{W}_u \in \mathbb{R}^{d_h \times d_h}$, $\mathbf{W}_q \in \mathbb{R}^{d_h \times d_h}$, and $\mathbf{v}_a \in \mathbb{R}^{d_h}$ are all trainable attention parameters and $\mathbf{b}_q \in \mathbb{R}^{d_h}$ is a bias vector. $\alpha_t^i$ is the normalized attention score of the $i$th source word at the decoding time step $t$.

For our pointer network-based decoding model, we use three variants of the single attention model. First, we use $\mathbf{h}_{t-1}^D$ to calculate $\mathbf{q}_t^i$ in the attention mechanism. Next, we use $\mathbf{y}_{prev}$ to calculate $\mathbf{q}_t^i$, where $\mathbf{W}_q \in \mathbb{R}^{(8d_p + d_r) \times d_h}$. In the final variant, we obtain the attentive context vector by concatenating the two attentive vectors obtained using $\mathbf{h}_{t-1}^D$ and $\mathbf{y}_{prev}$. This gives the best performance with the pointer network-based decoding model. These variants are referred to as $\mathrm{dec_{hid}}$, $\mathrm{tup_{prev}}$, and $\mathrm{combo}$ in Table \ref{tab:ablation}.

\subsection{Loss Function}

We minimize the negative log-likelihood loss of the generated words for word-level decoding ($\mathcal{L}_{word}$) and minimize the sum of negative log-likelihood loss of relation classification and the four pointer locations for pointer network-based decoding ($\mathcal{L}_{ptr}$).
\begin{align*}
&\mathcal{L}_{word} = -\frac{1}{B \times T} \sum_{b=1}^{B} \sum_{t=1}^{T} \text{log} (v_{t}^b)\\
&\mathcal{L}_{ptr} = -\frac{1}{B \times T} \sum_{b=1}^{B} \sum_{t=1}^{T} [ \text{log} (r_{t}^b) + \sum_{c=1}^{2} \text{log} (s_{c,t}^{b} e_{c,t}^{b})]
\end{align*}
 \noindent $v_t^b$ is the softmax score of the target word at time step $t$ for the word-level decoding model. $r$, $s$, and $e$ are the softmax score of the corresponding true relation label, true start and end pointer location of an entity. $b$, $t$, and $c$ refer to the $b$th training instance, $t$th time step of decoding, and the two entities of a tuple respectively. $B$ and $T$ are the batch size and maximum time step of the decoder respectively.

\section{Experiments}

\subsection{Datasets}

We focus on the task of extracting multiple tuples with overlapping entities from sentences. We choose the New York Times (NYT) corpus for our experiments. This corpus has multiple versions, and we choose the following two versions as their test dataset has significantly larger number of instances of multiple relation tuples with overlapping entities. (i) The first version is used by \citeauthor{zeng2018copyre} (\citeyear{zeng2018copyre}) (mentioned as NYT in their paper) and has $24$ relations. We name this version as NYT24. (ii) The second version is used by \citeauthor{hrlre@takanobu} (\citeyear{hrlre@takanobu}) (mentioned as NYT10 in their paper) and has $29$ relations. We name this version as NYT29. We select 10\% of the original training data and use it as the validation dataset. The remaining 90\% is used for training. We include statistics of the training and test datasets in Table \ref{tab:data_stat}. 

\begin{table}[ht]
\centering
\resizebox{0.85\columnwidth}{!}{
\begin{tabular}{lcc|cc}
\hline
       & \multicolumn{2}{c|}{NYT29} & \multicolumn{2}{c}{NYT24} \\ 
       & Train            & Test           & Train               & Test               \\ \hline
\# relations  & 29           & 29          & 24              & 24              \\ 
\# sentences    & 63,306           & 4,006          & 56,196              & 5,000              \\ 
\# tuples     & 78,973           & 5,859          & 88,366              & 8,120 \\ \hline
Entity overlap type       &             &            & \\ 
NEO  & 53,444           & 2,963          & 37,371              & 3,289              \\ 
EPO     & 8,379           & 898          & 15,124              & 1,410              \\ 
SEO    & 9,862           & 1,043          & 18,825              & 1,711              \\ \hline
\# tuples in a sentence       &             &            & \\ 
1 & 53,001           & 2,950          & 36,835              & 3,240              \\ 
2    & 6,154           & 595          & 12,065              & 1,047              \\ 
3    & 3,394           & 187          & 3,672              & 314              \\ 
4    & 450           & 239          & 2,623              & 290              \\ 
$\geq 5$    & 307           & 35          & 1,001              & 109              \\ \hline
\end{tabular}
}
\caption{Statistics of train/test split of the two datasets.}
\label{tab:data_stat}
\end{table}

\subsection{Parameter Settings}

We run the Word2Vec \cite{mikolov2013distributed} tool on the NYT corpus to initialize the word embeddings. The character embeddings and relation embeddings are initialized randomly. All embeddings are updated during training. We set the word embedding dimension $d_w=300$, relation embedding dimension $d_r=300$, character embedding dimension $d_c=50$, and character-based word feature dimension $d_f=50$. To extract the character-based word feature vector, we set the CNN filter width at $3$ and the maximum length of a word at $10$. The hidden dimension $d_h$ of the decoder LSTM cell is set at $300$ and the hidden dimension of the forward and the backward LSTM of the encoder is set at $150$. The hidden dimension of the forward and backward LSTM of the pointer networks is set at $d_p=300$. The model is trained with mini-batch size of $32$ and the network parameters are optimized using Adam \cite{kingma2014adam}. Dropout layers with a dropout rate fixed at $0.3$ are used in our network to avoid overfitting. 

\subsection{Baselines and Evaluation Metrics}

We compare our model with the following state-of-the-art joint entity and relation extraction models:

\noindent(1) {\em SPTree} \cite{miwa2016end}: This is an end-to-end neural entity and relation extraction model using sequence LSTM and Tree LSTM. Sequence LSTM is used to identify all the entities first and then Tree LSTM is used to find the relation between all pairs of entities.

\noindent(2) {\em Tagging} \cite{zheng2017joint}: This is a neural sequence tagging model which jointly extracts the entities and relations using an LSTM encoder and an LSTM decoder. They used a Cartesian product of entity tags and relation tags to encode the entity and relation information together. This model does not work when tuples have overlapping entities.

\noindent(3) {\em CopyR} \cite{zeng2018copyre}: This model uses an encoder-decoder approach for joint extraction of entities and relations. It copies only the last token of an entity from the source sentence. Their best performing multi-decoder model is trained with a fixed number of decoders where each decoder extracts one tuple. 

\noindent(4) {\em HRL} \cite{hrlre@takanobu}: This model uses a reinforcement learning (RL) algorithm with two levels of hierarchy for tuple extraction. A high-level RL finds the relation and a low-level RL identifies the two entities using a sequence tagging approach. This sequence tagging approach cannot always ensure extraction of exactly two entities.

\noindent(5) {\em GraphR} \cite{fu-etal-2019-graphrel}: This model considers each token in a sentence as a node in a graph, and edges connecting the nodes as relations between them. They use graph convolution network (GCN) to predict the relations of every edge and then filter out some of the relations.

\noindent(6) {\em N-gram Attention} \cite{trisedya-etal-2019-neural}: This model uses an encoder-decoder approach with N-gram attention mechanism for knowledge-base completion using distantly supervised data. The encoder uses the source tokens as its vocabulary and the decoder uses the entire Wikidata \cite{wikidata} entity IDs and relation IDs as its vocabulary. The encoder takes the source sentence as input and the decoder outputs the two entity IDs and relation ID for every tuple. During training, it uses the mapping of entity names and their Wikidata IDs of the entire Wikidata for proper alignment. Our task of extracting relation tuples with the raw entity names from a sentence is more challenging since entity names are not of fixed length. Our more generic approach is also helpful for extracting new entities which are not present in the existing knowledge bases such as Wikidata. We use their N-gram attention mechanism in our model to compare its performance with other attention models (Table \ref{tab:ablation}).

We use the same evaluation method used by \citeauthor{hrlre@takanobu} (\citeyear{hrlre@takanobu}) in their experiments. We consider the extracted tuples as a set and remove the duplicate tuples. An extracted tuple is considered as correct if the corresponding full entity names are correct and the relation is also correct. We report precision, recall, and F1 score for comparison. 

\subsection{Experimental Results}

Among the baselines, HRL achieves significantly higher F1 scores on the two datasets. We run their model and our models five times and report the median results in Table \ref{tab:comparison}. Scores of other baselines in Table \ref{tab:comparison} are taken from previous published papers \cite{zeng2018copyre,hrlre@takanobu,fu-etal-2019-graphrel}. Our {\em WordDecoding} (WDec) model achieves F1 scores that are $3.9\%$ and $4.1\%$ higher than HRL on the NYT29 and NYT24 datasets respectively. Similarly, our {\em PtrNetDecoding} (PNDec) model achieves F1 scores that are $3.0\%$ and $1.3\%$ higher than HRL on the NYT29 and NYT24 datasets respectively. We perform a statistical significance test (t-test) under a bootstrap pairing between HRL and our models and see that the higher F1 scores achieved by our models are statistically significant ($p < 0.001$). Next, we combine the outputs of five runs of our models and five runs of HRL to build ensemble models. For a test instance, we include those tuples which are extracted in the majority ($\geq 3$) of the five runs. This ensemble mechanism increases the precision significantly on both datasets with a small improvement in recall as well. In the ensemble scenario, compared to HRL, WDec achieves $4.2\%$ and $3.5\%$ higher F1 scores and PNDec achieves $4.2\%$ and $2.9\%$ higher F1 scores on the NYT29 and NYT24 datasets respectively.

\begin{table}[ht]
\centering
\resizebox{0.9\columnwidth}{!}{
\begin{tabular}{lllllll}
\hline
 & \multicolumn{3}{c}{NYT29} & \multicolumn{3}{c}{NYT24} \\ 
Model & Prec. & Rec. & F1 & Prec. & Rec. & F1 \\ \hline
Single   \\
Tagging & 0.593 & 0.381 & 0.464  & 0.624    & 0.317    & 0.420  \\ 
CopyR & 0.569 & 0.452 & 0.504  & 0.610    & 0.566    & 0.587   \\ 
SPTree & 0.492 & 0.557 & 0.522  &- &- &-  \\ 
GraphR & - & - & -  & 0.639 & 0.600 & 0.619  \\ 
HRL & 0.692 & 0.601 & 0.643  & 0.781    & 0.771    & 0.776  \\ 
WDec & \textbf{0.777} & 0.608 & \textbf{0.682}  & \textbf{0.881} & 0.761 & \textbf{0.817}  \\ 
PNDec & 0.732 & \textbf{0.624} & 0.673 & 0.806 & \textbf{0.773} & 0.789  \\ \hline
Ensemble  \\
HRL & 0.764 & 0.604 & 0.674 & 0.842 & 0.778 & 0.809  \\ 
WDec & \textbf{0.846} & 0.621 & \textbf{0.716} & \textbf{0.945} & 0.762 & \textbf{0.844}  \\ 
PNDec & 0.815 & \textbf{0.639} & \textbf{0.716} & 0.893 & \textbf{0.788} & 0.838    \\ \hline
\end{tabular}
}
\caption{Performance comparison on the two datasets.}
\label{tab:comparison}
\end{table}

\section{Analysis and Discussion}

\subsection{Ablation Studies}

We include the performance of different attention mechanisms with our {\em WordDecoding} model, effects of our masking-based copy mechanism, and ablation results of three variants of the single attention mechanism with our {\em PtrNetDecoding} model in Table \ref{tab:ablation}. {\em WordDecoding} with single attention achieves the highest F1 score on both datasets. We also see that our copy mechanism improves F1 scores by around 4--7\% in each attention mechanism with both datasets. {\em PtrNetDecoding} achieves the highest F1 scores when we combine the two attention mechanisms with respect to the previous hidden vector of the decoder LSTM ($\mathbf{h}_{t-1}^D$) and representation of all previously extracted tuples ($\mathbf{y}_{prev}$). 

\begin{table}[ht]
\centering
\resizebox{0.9\columnwidth}{!}{
\begin{tabular}{lllllll}
\hline
 & \multicolumn{3}{c}{NYT29} & \multicolumn{3}{c}{NYT24} \\ 
Model & Prec. & Rec. & F1 & Prec. & Rec. & F1 \\ \hline
WDec \\
Avg. & 0.638 & 0.523 & 0.575 & 0.771 & 0.683 & 0.724  \\ 
+ copy & 0.709 & 0.561 & 0.626 & 0.843 & 0.717 & 0.775  \\ 
N-gram & 0.640 & 0.498 & 0.560 & 0.783 & 0.698 & 0.738  \\ 
+ copy & 0.739 & 0.519 & 0.610 & 0.847 & 0.716 & 0.776  \\ 
Single & 0.683 & 0.545 & 0.607 & 0.816 & 0.716 & 0.763  \\ 
+ copy & \textbf{0.777} & \textbf{0.608} & \textbf{0.682} & \textbf{0.881} & \textbf{0.761} & \textbf{0.817}  \\ \hline
PNDec \\ 
$\mathrm{dec_{hid}}$ & 0.720 & 0.615 & 0.663 & 0.798 & 0.772 & 0.785  \\ 
$\mathrm{tup_{prev}}$ & 0.726 & 0.614 & 0.665 & 0.805 & 0.764 & 0.784  \\ 
$\mathrm{combo}$ & \textbf{0.732} & \textbf{0.624} & \textbf{0.673} & \textbf{0.806} & \textbf{0.773} & \textbf{0.789}  \\ \hline
\end{tabular}
}
\caption{Ablation of attention mechanisms with {\em WordDecoding} (WDec) and {\em PtrNetDecoding} (PNDec) model.}
\label{tab:ablation}
\end{table}

\subsection{Performance Analysis}

From Table \ref{tab:comparison}, we see that CopyR, HRL, and our models achieve significantly higher F1 scores on the NYT24 dataset than the NYT29 dataset. Both datasets have a similar set of relations and similar texts (NYT). So task-wise both datasets should pose a similar challenge. However, the F1 scores suggest that the NYT24 dataset is easier than NYT29. The reason is that NYT24 has around 72.0\% of overlapping tuples between the training and test data (\% of test tuples that appear in the training data with different source sentences). In contrast, NYT29 has only 41.7\% of overlapping tuples. Due to the memorization power of deep neural networks, it can achieve much higher F1 score on NYT24. The difference between the F1 scores of {\em WordDecoding} and {\em PtrNetDecoding} on NYT24 is marginally higher than NYT29, since {\em WordDecoding} has more trainable parameters (about 27 million) than {\em PtrNetDecoding} (about 24.5 million) and NYT24 has very high tuple overlap. However, their ensemble versions achieve closer F1 scores on both datasets.

Despite achieving marginally lower F1 scores, the pointer network-based model can be considered more intuitive and suitable for this task. {\em WordDecoding} may not extract the special tokens and relation tokens at the right time steps, which is critical for finding the tuples from the generated sequence of words. {\em PtrNetDecoding} always extracts two entities of varying length and a relation for every tuple. We also observe that {\em PtrNetDecoding} is more than two times faster and takes one-third of the GPU memory of {\em WordDecoding} during training and inference. This speedup and smaller memory consumption are achieved due to the fewer number of decoding steps of {\em PtrNetDecoding} compared to {\em WordDecoding}. {\em PtrNetDecoding} extracts an entire tuple at each time step, whereas {\em WordDecoding} extracts just one word at each time step and so requires eight time steps on average to extract a tuple (assuming that the average length of an entity is two). The softmax operation at the projection layer of {\em WordDecoding} is applied across the entire vocabulary and the vocabulary size can be large (more than 40,000 for our datasets). In case of {\em PtrNetDecoding}, the softmax operation is applied across the sentence length (maximum of 100 in our experiments) and across the relation set (24 and 29 for our datasets). The costly softmax operation and the higher number of decoding time steps significantly increase the training and inference time for {\em WordDecoding}. The encoder-decoder model proposed by \citeauthor{trisedya-etal-2019-neural} (\citeyear{trisedya-etal-2019-neural}) faces a similar softmax-related problem as their target vocabulary contains the entire Wikidata entity IDs and relation IDs which is in the millions. HRL, which uses a deep reinforcement learning algorithm, takes around 8x more time to train than {\em PtrNetDecoding} with a similar GPU configuration. The speedup and smaller memory consumption will be useful when we move from sentence-level extraction to document-level extraction, since document length is much higher than sentence length and a document contains a higher number of tuples.

\subsection{Error Analysis}

The relation tuples extracted by a joint model can be erroneous for multiple reasons such as: (i) extracted entities are wrong; (ii) extracted relations are wrong; (iii) pairings of entities with relations are wrong. To see the effects of the first two reasons, we analyze the performance of HRL and our models on entity generation and relation generation separately. For entity generation, we only consider those entities which are part of some tuple. For relation generation, we only consider the relations of the tuples. We include the performance of our two models and HRL on entity generation and relation generation in Table \ref{tab:ent_rel}. Our proposed models perform better than HRL on both tasks. Comparing our two models, {\em PtrNetDecoding} performs better than {\em WordDecoding} on both tasks, although {\em WordDecoding} achieves higher F1 scores in tuple extraction. This suggests that {\em PtrNetDecoding} makes more errors while pairing the entities with relations. We further analyze the outputs of our models and HRL to determine the errors due to ordering of entities (Order), mismatch of the first entity (Ent1), and mismatch of the second entity (Ent2) in Table \ref{tab:error}. {\em WordDecoding} generates fewer errors than the other two models in all the categories and thus achieves the highest F1 scores on both datasets.

\begin{table}[ht]
\centering
\resizebox{0.8\columnwidth}{!}{
\begin{tabular}{l|l|l|lll}
\hline
                       &                           & Model      & Prec. & Rec.  & F1    \\ \hline
\multirow{6}{*}{NYT29} & \multirow{3}{*}{Ent}   & HRL      & 0.833 & 0.827 & 0.830 \\
                       &                             & WDec      & \textbf{0.865} & 0.812 & 0.838 \\
                       &                           & PNDec & 0.858 & \textbf{0.851} & \textbf{0.855} \\ \cline{2-6} 
                       & \multirow{3}{*}{Rel} & HRL      & 0.846 & 0.745 & 0.793 \\  
                       &                         & WDec      & \textbf{0.895} & 0.729 & 0.803 \\
                       &                           & PNDec & 0.884 & \textbf{0.770} & \textbf{0.823} \\ \hline
\multirow{6}{*}{NYT24} & \multirow{3}{*}{Ent}   & HRL      & 0.887 & 0.892 & 0.890 \\
                       &                        & WDec      & \textbf{0.926} & 0.858 & 0.891 \\
                       &                           & PNDec & 0.906 & \textbf{0.901} & \textbf{0.903} \\ \cline{2-6} 
                       & \multirow{3}{*}{Rel} & HRL      & 0.906 & 0.896 & 0.901 \\ 
                       &                        & WDec      & \textbf{0.941} & 0.880 & 0.909 \\
                       &                           & PNDec & 0.930 & \textbf{0.921} & \textbf{0.925} \\ \hline
\end{tabular}
}
\caption{Comparison on entity and relation generation tasks.}
\label{tab:ent_rel}
\end{table}

\begin{table}[ht]
\centering
\resizebox{0.8\columnwidth}{!}{
\begin{tabular}{lcccccc}
\hline
      & \multicolumn{3}{c}{NYT29} & \multicolumn{3}{c}{NYT24} \\ 
Model & Order    & Ent1   & Ent2   & Order    & Ent1   & Ent2   \\ \hline
HRL  & 0.2      & 5.9    & 6.6    & 0.2      & 4.7    & 6.3    \\
WDec  & 0.0      & 4.2    & 4.7    & 0.0      & 2.4    & 2.4    \\ 
PNDec & 0.8      & 5.6    & 6.0    & 1.0      & 4.0    & 6.1    \\ \hline
\end{tabular}
}
\caption{\% errors for wrong ordering and entity mismatch.}
\label{tab:error}
\end{table}

\section{Related Work}

Traditionally, researchers \cite{mintz2009distant,riedel2010modeling,hoffmann2011knowledge,zeng2014relation,zeng2015distant,huang2016attention,ren2017cotype,zhang-etal-2017-position,jat2018attention,vashishth2018reside,ye-ling-2019-distant,guo2019aggcn,nayak-ng-2019-effective} used a pipeline approach for relation tuple extraction where relations were identified using a classification network after all entities were detected. \citeauthor{su2018exploring} (\citeyear{su2018exploring}) used an encoder-decoder model to extract multiple relations present between two given entities. 

Recently, some researchers \cite{katiyar2016investigating,miwa2016end,bekoulis2018joint,dat2019end} tried to bring these two tasks closer together by sharing their parameters and optimizing them together. \citeauthor{zheng2017joint} (\citeyear{zheng2017joint}) used a sequence tagging scheme to jointly extract the entities and relations. \citeauthor{zeng2018copyre} (\citeyear{zeng2018copyre}) proposed an encoder-decoder model with copy mechanism to extract relation tuples with overlapping entities. \citeauthor{hrlre@takanobu} (\citeyear{hrlre@takanobu}) proposed a joint extraction model based on reinforcement learning (RL). \citeauthor{fu-etal-2019-graphrel} (\citeyear{fu-etal-2019-graphrel}) used a graph convolution network (GCN) where they treated each token in a sentence as a node in a graph and edges were considered as relations. \citeauthor{trisedya-etal-2019-neural} (\citeyear{trisedya-etal-2019-neural}) used an N-gram attention mechanism with an encoder-decoder model for completion of knowledge bases using distant supervised data.

Encoder-decoder models have been used for many NLP applications such as neural machine translation \cite{sutskever2014sequence,Bahdanau2014NeuralMT,luong2015effective}, sentence generation from structured data \cite{diego2018deep,trisedya-etal-2018-gtr}, and open information extraction \cite{zhang2017mt,neuralopenie}. Pointer networks \cite{vinyals2015pointer} have been used to extract a text span from text for tasks such as question answering \cite{seo2016bidirectional,kundu2017question}. For the first time, we use pointer networks with an encoder-decoder model to extract relation tuples from sentences.

\section{Conclusion}

Extracting relation tuples from sentences is a challenging task due to different length of entities, the presence of multiple tuples, and overlapping of entities among tuples. In this paper, we propose two {\em novel} approaches using encoder-decoder architecture to address this task. Experiments on the New York Times (NYT) corpus show that our proposed models achieve significantly improved new state-of-the-art F1 scores. As future work, we would like to explore our proposed models for a document-level tuple extraction task.

\section{Acknowledgments}

We would like to thank the anonymous reviewers for their valuable and constructive comments on this paper.

\begin{quote}
\begin{small}
\bibliographystyle{aaai}
\bibliography{10036_nayak_ng}
\end{small}
\end{quote}

\end{document}